\documentclass[11pt]{article}

\usepackage[preprint]{acl}

\usepackage{times}
\usepackage{latexsym}

\usepackage[T1]{fontenc}
\usepackage[utf8]{inputenc}

\usepackage{microtype}

\usepackage{inconsolata}

\usepackage{graphicx}
\usepackage{amsmath}

\title{Enhancing Japanese Large Language Models with Reasoning Vectors}

\author{
  \textbf{Carolina Minami Oguchi\textsuperscript{1}},
  \textbf{Leo Wei\textsuperscript{2,4}},
  \textbf{Koyo Kobayashi\textsuperscript{3}},
  \textbf{Hsin-Tai Wu\textsuperscript{4}},
  \textbf{Dipak Ghosal\textsuperscript{1}}
\\
  \textsuperscript{1}University of California, Davis,
  \textsuperscript{2}Santa Clara University, \\
  \textsuperscript{3}NTT DOCOMO, INC,
  \textsuperscript{4}DOCOMO Innovations, Inc
\\
  \small{
    \href{mailto:email@domain}{coguchi@ucdavis.edu}, 
    \href{mailto:email@domain}{twei2@scu.edu},
    \href{mailto:email@domain}{kouyou.kobayashi.gv@nttdocomo.com},
    \href{mailto:email@domain}{hwu@docomoinnovations.com},
    \href{mailto:email@domain}{dghosal@ucdavis.edu}
  }
}

\begin{document}
\maketitle
\begin{abstract}
Post-training methods have improved the performance and enhanced the reasoning capability for mainstream large language models (LLMs), but the same is challenging for Japanese LLMs to achieve due to the amount of resources required. Inspired by task vectors that extract the change of weights before and after training, specifically for a certain task, we obtain reasoning vectors from reasoning LLMs and apply them to Japanese LLMs to boost their performance. While the resources available present a challenge to improve Japanese LLMs, we present a simple and effective way to obtain high improvement and hope to inspire for other languages.
\end{abstract}

\section{Introduction}
Recent advancements in large language models (LLMs) have demonstrated that post-training techniques such as supervised fine-tuning (SFT) and reinforcement learning (RL) can substantially improve model performance. These methods depend heavily on human involvement for dataset curation and leverage major models for data quality filtering, which are primarily concentrated in English and other high-resource languages. The Qwen 3 flagship model \cite{qwen3} employs a four-stage post-training pipeline, utilizing Qwen2.5-72B-Instruct and QwQ-32B \cite{qwq} for query and response filtering, respectively. Human annotation is incorporated to resolve issues in response filtering. Similarly, s1-32B \cite{muennighoff2025s1} uses SFT to learn reasoning chains from a curated dataset of 1,000 samples, which excludes low-quality data through manual inspection and removes easy examples if they are correctly answered by Qwen 2.5 models. DeepSeek-R1 \cite{guo2025deepseek} engages human annotators to refine RL cold-start training data and leverages model checkpoints alongside DeepSeek-v3 \cite{liu2024deepseek} to generate and evaluate data for SFT.

In contrast, Japanese large language models (LLMs) face distinct challenges in applying post-training techniques. The limited availability of public datasets and expert annotators hinders the ability to conduct post-training at a comparable scale. Additionally, the lack of robust, large-scale Japanese models for effectively assessing and filtering data quality further complicates these efforts. Furthermore, reliance on machine-translated data from English can be problematic, as translations often fail to capture the linguistic nuances, cultural context, and full semantic richness of native Japanese, which may negatively affect model performance.

To overcome these challenges, we explore a novel approach that extracts reasoning vectors from mainstream LLMs and transfers them to Japanese models. By injecting these reasoning vectors, we aim to enhance the reasoning capabilities of Japanese LLMs without additional training.

\footnotetext[1]{https://huggingface.co/spaces/llm-jp/open-japanese-llm-leaderboard}

\begin{figure}[t]
  \includegraphics[width=\columnwidth]{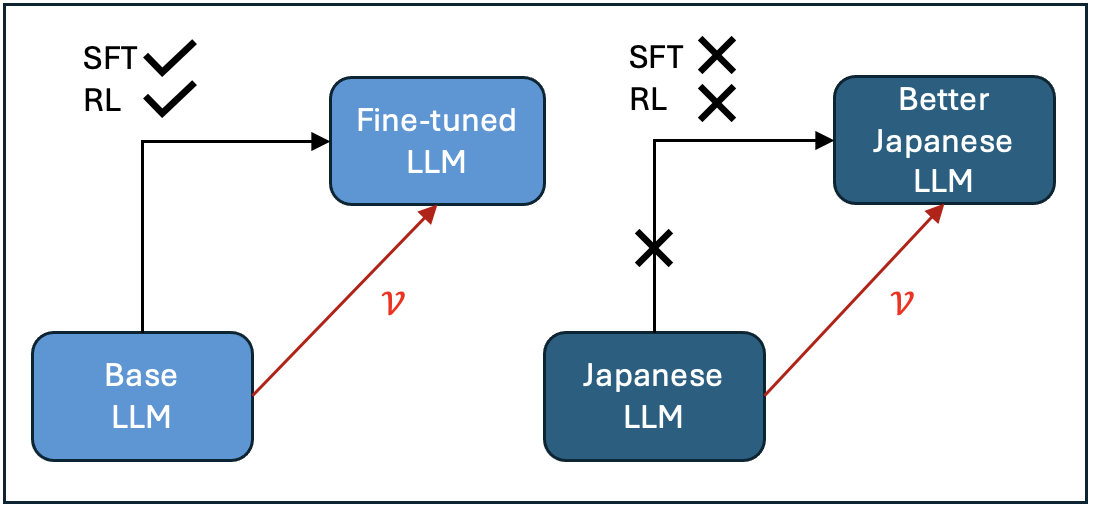}
  \caption{Unlike English LLMs, Japanese LLMs suffer from the amount of data to effectively scale up through supervised fine-tuning or reinforcement learning. By extracting a reasoning vector $v$ between the base and fine-tuned English LLMs and adding it to Japanese LLMs, they improve significantly on benchmark datasets.}
  \label{fig:intro}
\end{figure}

\section{Related Work}

\subsection{Japanese LLMs and Training Datasets}
\textbf{LLMs} While there have been a lot of models and training datasets for mainstream LLMs \cite{naveed2023comprehensive}, it is not the case for the Japanese domain. LLM-jp \cite{aizawa2024llm} presents a cross-organizational effort to conduct large-scope LLM training, including building training corpus from Japanese Wikipedia, incorporating instruction-following datasets, and adopting RL algorithms for post-training. Rinna \cite{sawada2024release} fine-tunes GPT on mostly Wikipedia and applies SFT and PPO \cite{schulman2017proximal} on post-training data translated from English. As another Japanese LLM, Swallow \cite{fujii2024continual} continues the pre-training stage of Llama2 \cite{touvron2023llama} on its corpus mainly sourced from Japanese Wikipedia.

\noindent \textbf{Datasets} For instruction-following datasets, \citet{hirano2023llm} contains 8 million samples on various tasks with the majority made from translating from English. \citet{suzuki2023base} refines this dataset by filtering for the non-translated portion and incorporating more samples from Japanese Wikipedia Typo Dataset \cite{tanaka2020building} and the Japanese Question-Answering Corpus \cite{tanioka2018automatic}. AnswerCarefully \cite{suzuki2025answercarefully} contains samples specifically for promoting model safety.

\subsection{Task Vector}
\citet{ilharcoediting} shows that the difference in model weights between the two models presents a meaningful direction between the two. They propose task vectors to set this direction as learning a specific task and find that adding this vector to target models can enhance their performance on the task, without any additional training. Chat vector \cite{huang2024chat} adopts the same concept by subtracting the weights between a pre-trained and an instruction fine-tuned model, enabling LLMs to chat in other languages. The approach first fine-tunes the LLM on Traditional Chinese corpus through continual pre-training to obtain a Chinese LLM. The chat vector is added to the LLM to equip it with instruction-following abilities without fine-tuning on relevant datasets. \citet{wu2025unlocking} applies task vector to reduce the output sequence length of reasoning models.

\section{Reasoning Vector}
Given a pre-trained model $\pi_{\text{pre}}$ and a post-trained model $\pi_{\text{post}}$, the reasoning vector $v$ is obtained by subtracting the weights between the two models:

\begin{equation}
    v = \pi_{\text{post}} - \pi_{\text{pre}}
\end{equation}

\noindent Given a target model $\pi_{\text{tgt}}$, the reasoning vector $v$ is subsequently added to $\pi_{\text{tgt}}$ with a scalar weight $w$ to obtain the enhanced model $\pi_{\text{enh}}$:

\begin{equation}
    \pi_{\text{enh}} = \pi_{\text{tgt}} + w * v
\end{equation}
    
\noindent The vector represents the direction of post-training in the weight space and adding this vector to the model equips it with post-training knowledge.

\section{Experiments}

\subsection{Setup}
We adopt Qwen-32B as the pre-trained model $\pi_{\text{pre}}$, s1-32B as the post-trained model $\pi_{\text{post}}$, and EZO\footnote{https://huggingface.co/AXCXEPT/EZO-Qwen2.5-32B-Instruct} as the target model $\pi_{\text{tgt}}$. s1-32B learned to reason through SFT on reasoning chains, and EZO underwent instruction pre-training on Japanese data.
These models were selected because they represent the only set for which we obtained all pre-trained, post-trained, and target models that share the same architecture.

%\subsection{Baselines}
%For baseline models, we select mainstream models and Japanese-specific models from the leaderboard. Mainstream models include Llama 3, \cite{grattafiori2024llama}, Qwen 2.5 \cite{qwen2.5}, and DeepSeek-R1 \cite{guo2025deepseek}. For models fine-tuned for Japanese capability, Youko \cite{sawada2024release} undergoes SFT and gains the chat capability through obtaining the chat vector \cite{huang2024chat} between Llama 3 and Llama 3-Instruct. LLM-jp \cite{aizawa2024llm} continues the pre-training stage with instruction pre-training \cite{cheng2024instruction} and adopts DPO \cite{rafailov2023direct} for RL post-training. Please refer to the leaderboard\footnotemark[2] for other baseline models. 

\subsection{Evaluation Datasets}
%\textbf{Japanese Leaderboard}\footnotemark[2] The leaderboard includes dozens of datasets spanning various domains including natural language inference, question answering and mathematical reasoning, and we adopt its evaluation protocol and metrics.

%\footnotetext[2]{https://huggingface.co/spaces/llm-jp/open-japanese-llm-leaderboard}
%\setcounter{footnote}{2}

%\noindent \textbf{MATH500}\footnote{https://huggingface.co/datasets/appier-ai-research/Multilingual-MATH-500} The dataset \cite{hendrycks2measuring} contains 500 competition-level mathematical problems selected by prior work \cite{lightman2023let}.

\noindent \textbf{AIME24}\footnote{https://artofproblemsolving.com/wiki/index.php/2024\_AIME\_I} The American Invitational Mathematics Examination contains 30 challenging problems to select talented high school students for the Olympiad competition. This dataset is translated into Japanese by manually elaborating ChatGPT's translation.

\subsection{Grading}
Common approaches to grading the answers include extracting the final numerical value and comparing it to the ground-truth label, or using a separate language model to assess correctness. In this work, we adopt a simpler heuristic: a response is considered correct if it contains the answer label. This method was chosen for its ease of implementation and serves as a temporary solution. A more robust evaluation strategy will be incorporated in future work.

\section{Experimental Results and Discussion}

\begin{figure}[t]
  \includegraphics[width=\columnwidth]{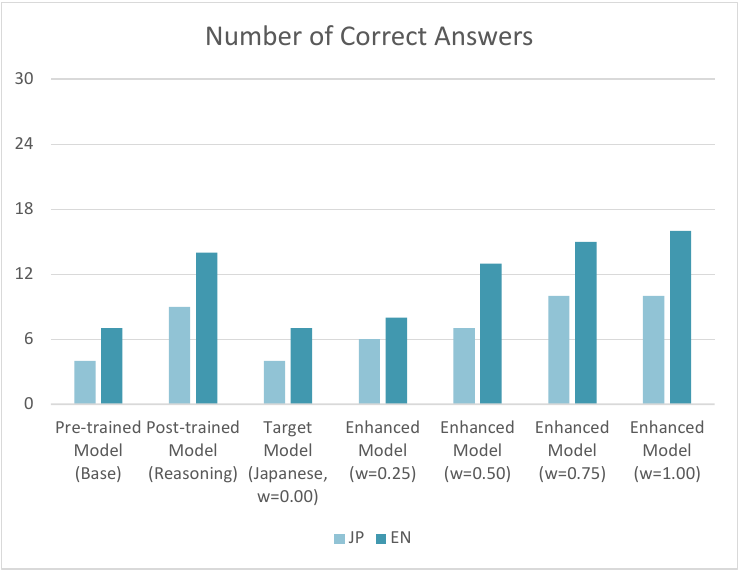}
  \caption{Number of correct answers on Japanese and English AIME 2024 problems for each model}
  \label{fig:aime_results}
\end{figure}

Table~\ref{tab:aime_results} shows the number of correct answers that each model achieved on the original AIME24 dataset and the translated Japanese AIME24 dataset, based on a single evaluation run. It is visualized in Figure~\ref{fig:aime_results}.
%don't need both table and graph. just a graph would be enough?

\begin{table}[]
    \centering
    \begin{tabular}{|l||l|l|}
        \hline
        Model                             & JP & EN \\ \hline \hline
        Pre-trained Model (Base)          & 4  & 7  \\ \hline
        Post-trained Model (Reasoning)    & 9  & 14 \\ \hline
        Target Model (Japanese, $w=0.00$) & 4  & 7  \\ \hline
        Enhanced Model   $w=0.25$   & 6  & 8  \\ \hline
        Enhanced Model   $w=0.50$   & 7  & 13 \\ \hline
        Enhanced Model   $w=0.75$   & 10 & 15 \\ \hline
        Enhanced Model   $w=1.00$   & 10 & 16 \\ \hline
    \end{tabular}
    \caption{Number of correct answers out of $30$ questions in AIME24}
    \label{tab:aime_results}
\end{table}

%Figure~\ref{fig:aime_results} shows how well each model performed on the Japanese AIME24 dataset and the original English AIME24 dataset.
Among the given models, the pre-trained model has the minimum performance, the post-trained model has a better performance, and the target model has the minimum performance too. Incorporating the reasoning vector into the target model with varying weights ($w$) leads to consistent performance gains. At $w = 0.25$, the enhanced model slightly outperforms both the pre-trained and target models. As $w$ increases to $0.50$, performance improves further. Notably, for higher values of $w$, the enhanced model surpasses the original post-trained reasoning model.

These results highlight the effectiveness of reasoning vector integration in improving task performance. As the reasoning vector is incrementally incorporated into the target model, its performance consistently improves. This suggests that the enhanced model successfully acquires reasoning capabilities from the reasoning vector.

\section{Conclusions and Future Work}
This work demonstrates that the reasoning capabilities of large language models can be effectively transferred to Japanese models through reasoning vectors. By extracting the difference in weights between a pre-trained and post-trained model trained for reasoning, we successfully enhanced a Japanese instruction-tuned target model without requiring any additional training or labeled data. Our experiments on the AIME24 dataset reveal that injecting reasoning vectors consistently improves performance, even surpassing the original reasoning model at higher weights.

This simple yet effective approach offers a promising path for boosting under-resourced language models using existing advancements in other languages. It opens new avenues for leveraging post-training knowledge when direct training is infeasible.

Future work includes applying this approach to smaller models and quantized models to explore its feasibility. 

\section*{Limitations}
While the reasoning vector enhances Japanese models without additional fine-tuning and data curation, we present two limitations. First, the pre-trained, post-trained, and target models have to follow the same model architecture for elementwise arithmetic operations to work. Second, for evaluation datasets without a training set, it lacks a heldout set to determine the weight of the reasoning vector when adding to the target model.
%\section*{Acknowledgments}

\bibliography{main}

%\appendix

%\section{Example Appendix}
%\label{sec:appendix}

%This is an appendix.

\end{document}